\documentclass[10pt,journal,compsoc]{IEEEtran}

\usepackage{cite}
\usepackage{comment}
\usepackage{xspace}
\usepackage{multirow}
\usepackage{algorithm}
\usepackage{amsmath,amsfonts}

\usepackage{cite}
\usepackage{amsmath,amssymb,amsfonts}
\usepackage{algorithm}
\usepackage{algorithmic}
\usepackage{url}
\usepackage{comment}
\usepackage{graphicx}
\usepackage[square,numbers,sort&compress]{natbib}
\usepackage{textcomp}
\usepackage{soul}
\usepackage{xcolor}
\usepackage{multirow}

\newcommand{\ie}{{\itshape i.e.}\xspace}
\newcommand{\eg}{\emph{e.g.}\xspace}
\newcommand{\etc}{\emph{etc.}\xspace}

\newcommand{\eph}[1]{\textbf{\textit{#1}}}
\makeatletter
\font\myfont=cmr12 at 21pt

\begin{document}

\title{\myfont A Survey of Large-Scale Deep Learning Serving System Optimization: Challenges and Opportunities}

\author{Fuxun Yu{$^\dagger$}, Di Wang{$^\ddagger$}, Longfei Shangguan{$^\ddagger$}, Minjia Zhang{$^\ddagger$}, Xulong Tang{$^*$}, Chenchen Liu$^\S$, Xiang Chen{$^\dagger$}
\thanks{{$^\dagger$}George Mason University, {$^\dagger$}Microsoft,
{$^\dagger$}University of Pittsburgh, {$^\S$}University of Maryland, Baltimore County}}

\maketitle

\begin{abstract}
  Deep Learning (DL) models have achieved superior performance in many application domains, including vision, language, medical, commercial ads, entertainment, etc.
  With the fast development, both DL applications and the underlying serving hardware have demonstrated strong scaling trends, i.e., \textit{Model Scaling} and \textit{Compute Scaling}, for example, the recent pre-trained model with hundreds of billions of parameters with $\sim$TB level memory consumption, as well as the newest GPU accelerators providing hundreds of TFLOPS. 
  With both scaling trends, new problems and challenges emerge in DL inference serving systems, which gradually trends towards \textit{Large-scale Deep learning Serving system (LDS)}.
  This survey aims to summarize and categorize the emerging challenges and optimization opportunities for large-scale deep learning serving systems. By providing a novel taxonomy, summarizing the computing paradigms, and elaborating the recent technique advances, we hope that this survey could shed lights on new optimization perspectives and motivate novel works in large-scale deep learning system optimization.  
\end{abstract}

\section{Introduction}

\IEEEPARstart{D}{eep} Learning (DL) models, such as CNNs~\cite{r50, mbnetv2, mnasnet}, transformers~\cite{vit, bert, gpt, mega-turing} and recommendation models~\cite{dlrm, dlrm_survey} have achieved superior performance in many cognitive tasks like vision, speech and language applications, which poses potentials in numerous areas like medical image analysis~\cite{medical}, photo styling~\cite{guagan}, machine translation~\cite{trans}, product recommendation~\cite{dlrm}, customized advertising~\cite{ads}, game playing~\cite{game}, \etc 
Such widespread DL applications bring great market values and lead to significant DL serving traffics. 
For example, FB has 1.82 billions of daily active users~\cite{fb_busi}. The number of advertising recommendation queries can reach 10M queries per second. 
The huge growth in consumer-generated data and use of DL services have also propelled the demand for AI-centric data centers (such as Amazon AWS~\cite{aws} and Microsoft Azure~\cite{azure}) and the increased adoption of powerful DL accelerators like GPUs.
According to the report~\cite{gpu_report}, GPU has accounted for the major share of 85\% with 2983M USD in the global data center accelerator market in 2018. And this product segment is poised to reach 29819M USD by 2025~\cite{gpu_report}.

\begin{figure*}[!t]
	\centering
	\includegraphics[width=6.7in]{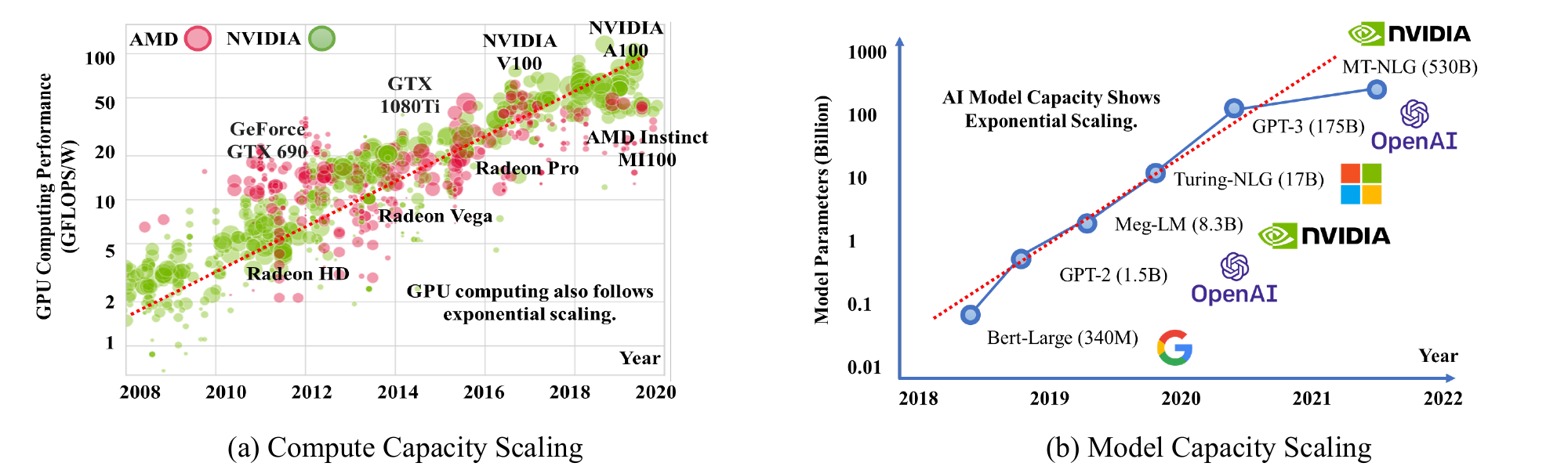}
	\caption{Deep Learning {Model Scaling} and {Computing Scaling}~\cite{yifan}. Such exponential scaling trends on both sides leads to {Large-scale DL Systems}.}
	\vspace{-2mm}
	\label{fig:scaling}
\end{figure*}

With the ever increasing market demands, DL applications and the underlying serving hardware have demonstrate strong {scaling trends} in terms of \textbf{Computing Scaling} (\eg, increased computing parallelism, memory and storage to serve larger models) and \textbf{Model Scaling} (\eg, higher structure complexity, computing workload, parameter size for better accuracy), which greatly complicates the serving system management and optimization.
%
%
%
On the one hand, as shown in Figure~\ref{fig:scaling} (a), with the \textbf{Computing Scaling} trend, GPU with massive computing parallelism has become one of the major types of DL computing accelerators in recent data centers and maintains continuously exponential performance scaling.
Recent GPUs such as NVIDIA Tesla V100 offer 130 Tera floating point operations per second (TFLOPS), and 900 GB/s memory bandwidth, and these numbers further increase to 312 TFLOPS and 1.6TB/s memory bandwidth, which can serve tens of DL models such as ResNet50~\cite{r50} simultaneously and provide higher efficiency (Perf/Watt). 
On the other hand, as shown in Figure~\ref{fig:scaling} (b), \textbf{Model Scaling} has been show to be one of the most important factor in achieving better accuracy, the effectiveness of which is consistently shown in practice by industrial ultra-large models in all domains, such as vision model BiT~\cite{bit}, NLP model BERT~\cite{bert}, GPT3~\cite{gpt} and deep learning recommendation model DLRM~\cite{dlrm}.
For example, recent ultra-large model MT-NLG~\cite{mega-turing} has achieved 530 billions of parameters. Industrial-level commercial DLRMs~\cite{dlrm} have reached $\sim$TB model size, which significantly surpass single-machine memory capability and require multiple devices for collaborative computing.

{{Within such contexts, we observe the current DL system community still lacks enough awareness and attention to such \textbf{Large-scale Deep Learning Systems (LDS)},  overlooking the emerged challenges and opportunities:}}
    Traditional DL system optimization usually focuses on the single-model single-machine inference setting (\ie, one-to-one mapping).  However, LDS with bigger DL models and more powerful hardwares enable more flexible inference computing, bringing the many-instance to one-device, one-instance to many-device, and even many-instance to many-device mapping into practice.
For example, computing scaling (such as GPUs, TPUs) motivates many research works into multi-model inference on one single device, \eg, splitting one GPU into multiple containerized vGPUs or Multi-Instance GPUs (MIG) for better hardware utilization, higher serving throughput and cost efficiency. 
    Considering practical cost management (\eg, total cost of ownership, TCO), data centers serving massive inference queries also tend to migrate to {multi-tenant} inference serving, \eg, co-locating multiple inference queries on the same device, which incurs new optimization objectives (\eg, total served queries per second, QPS) and constraints (\eg, service-level agreement, SLA) from traditional single-tenant inference.
Similarly, the model scaling also poses requests for new one-to-many inference scenarios. The recent ultra-large model (\eg, DLRMs) incurs huge memory cost ($\sim$TB without quantization) during inference, which requires new collaborative computing paradigms such as heterogeneous computing or distributed inference. Such collaboratively serving involves remote process calls (RPCs) and low-bandwidth communications, which brings dramatically different bottlenecks from traditional single-device inference. 
    With all above scenarios involved, modern data centers face more sophisticated many-to-many scenarios and require dedicated inference query scheduling, such as service router and compute device management, for better serving performance like latency, throughput and cost, \etc

In this work, we propose a a novel computing paradigm taxonomy for ongoing LDS works, summarize the new optimization objectives, elaborate new technical design perspectives, and provide the insights for future LDS optimization.

\begin{itemize}
	\item \textbf{Multi-to-Multi Computing Paradigm} Characterized by the relations between DNN Instances (I) and Compute Devices (D), emerging LDS computing paradigms can be taxonomized into three new categories beyond Single Instance Single Device (\eph{SISD}), that is, Multi Instance Single Device (\eph{MISD}), Single Instance Multi Device (\eph{SIMD}) and Multi Instance Multi Devices (\eph{MIMD}), as shown in Figure~\ref{fig:taxonomy}.

	\vspace{1mm}
	\item \textbf{Inference Serving Oriented Objectives} Different from SISD that focuses on single-model performance, LDS works have different optimization goals, including inference \eph{latency, serving throughput, costs, scalability, quality of service}, \etc For example, multi-tenant inference (MISD) targets at improving the serving throughput and power efficiency, while super-scale model inference serving aims to enhance hardware scalability with low costs.

	\vspace{1mm}
	\item \textbf{At-Scale Design and Technologies} Due to the scale of inference serving, LDS works have also demonstrated many optimization challenges and opportunities within \eph{algorithm innovation}, \eph{runtime scheduling} and \eph{resource management}. For example, multi-tenant inference optimization seeks for fine-grained hardware resource partitioning and job scheduling, \eg, spatial/temporal sharing to provide QoS assurance.
	Distributed inference with slow communication bottlenecks requires dedicated model-hardware co-optimization, \eg, efficient model sharding and balanced collaboration, \etc

\end{itemize}

By summarizing the existing works, we aim to provide a comprehensive survey on emerging challenges, opportunities, and innovations, and thus motivates new innovations in LDS operation and optimization. 
	The rest of the survey is organized as follows:
	Section~\S\ref{sec:bg} presents the research preliminaries including our taxonomy for LDS and indicate the scope of this survey.
    Section~\S\ref{sec:misd} summarizes the challenges and recent works in multi-instance single-device (MISD) optimization;
    Section~\S\ref{sec:simd} summarizes the research works in single-instance multi-device (SIMD) optimization;
    Section~\S\ref{sec:con} concludes this work.

\section{Large-Scale DL Serving System: \\A Novel Taxonomy}
\label{sec:bg}

\begin{figure*}[!t]
	\centering
	\includegraphics[width=6.5in]{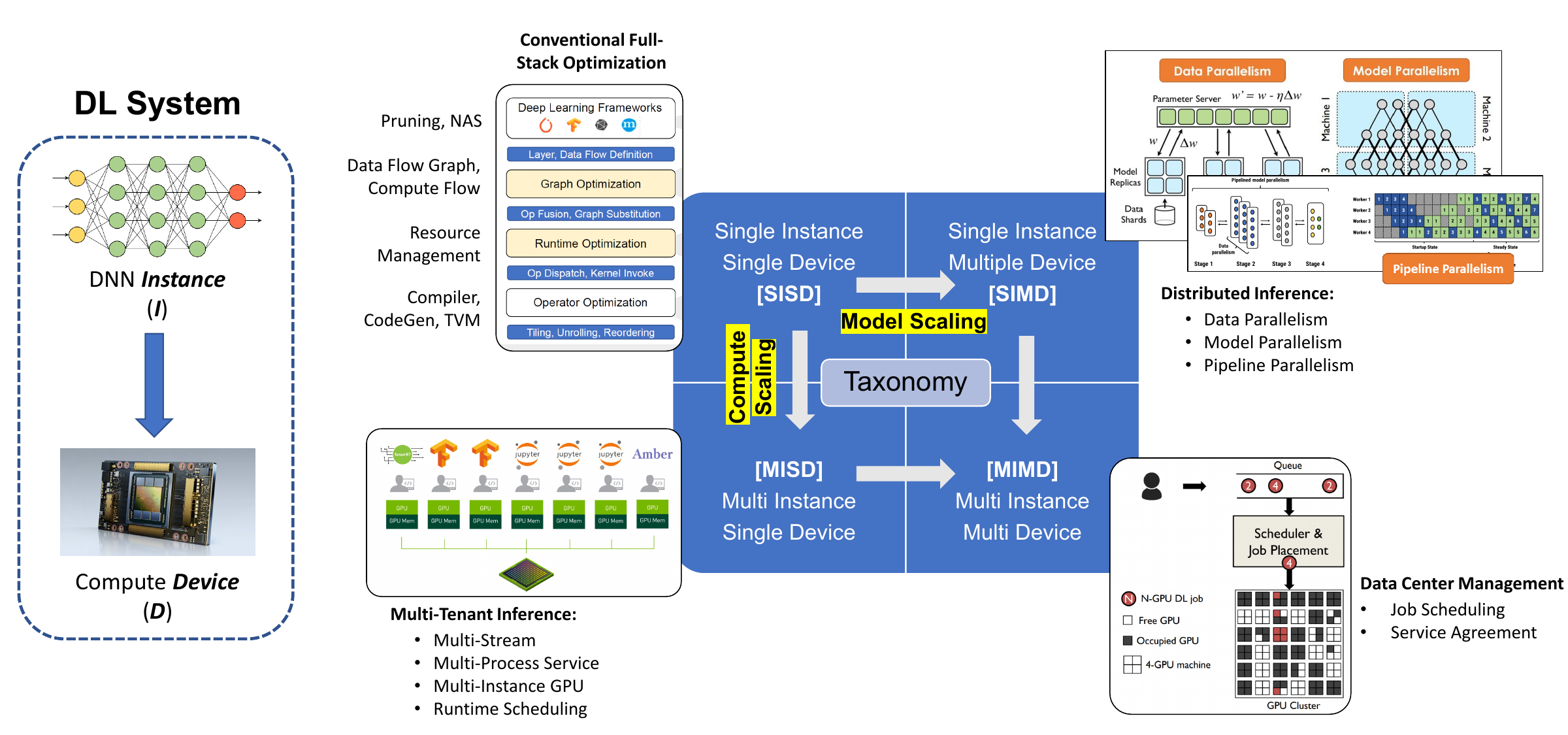}
	\caption{A Taxonomy of Deep Learning System from the relationship between DNN Instance \textbf{(I)} and Compute Device \textbf{(D)}: Single-Instance Single-Device, SISD, Single-Instance Multi-Device, SIMD, Multi-Instance Single-Device, MISD and Multi-Instance Multi-Device, MIMD.}
	\label{fig:taxonomy}
\end{figure*}

\vspace{1mm}
\noindent {\textbf{Taxonomy Overview}} We first give an high-level overview of LDS optimization taxonomy. 
Specifically, we use a taxonomy shown in Figure~\ref{fig:taxonomy} to demonstrate such differences between traditional DL system optimizations and new emerging challenges in LDS. 
We use \textit{Instance (I)} to denote one DNN model and \textit{Device (D)} to denote the underlying serving compute hardware.  
Traditional DL system optimizations, though incorporating a thorough full stack (\eg, with model-, graph-, runtime- and kernel-optimization levels), usually comes within the \textit{Single Instance Single Device (SISD)} assumption.
Therefore, most existing works only constitute the top-left quarter in the full-spectrum of DLS, neglecting the \textit{Single Instance Multiple Devices (SIMD)}, \textit{Multiple Instances Single Device (MISD)}, and even \textit{Multiple Instances Multiple Devices (MIMD)} optimizations.

\vspace{1mm}
\noindent $\bullet$ {\textbf{Single Instance Single Device (SISD)}} SISD optimization improves one model's end-to-end performance (such as latency) on the targeted hardware device, such as CPU, GPU, FPGA, \etc 
Conventional SISD optimizations have studied the full stacks thoroughly, including the algorithm-level NN design~\cite{mbnet,shufflenet, mobilenet}, pruning and NAS works~\cite{effnet,hrank,partial}, as well as the compiler-level optimization works~\cite{tvm, taso, cortex}.

Previous DL optimization is dominant by SISD works. For example, the popular DL compiler frameworks (\eg, TVM~\cite{tvm}, Ansor~\cite{ansor}, TASO~\cite{taso}, \etc) tune the low-level computing primitives to yield high-performance kernels for the given underlying hardware. 
However, with the fast evolving of DL models and AI hardwares, more applications and potentials come up in terms of MISD and SIMD optimization domains.
Although demonstrating optimal performance for SISD serving, these techniques are usually ill-suited for new scenarios multiple device co-serving one model (SIMD) or multi-model co-running on one device (MISD), which have distinct LDS-incurred computing challenges and opportunities. 


\vspace{1mm}
\noindent $\bullet$ {\textbf{Multi Instance Single Device (MISD)}} MISD optimization targets at improving the serving throughput (such as serving query per second, QPS) by co-locating multiple DNN instances on one high-performance hardware (\ie, multi-tenant inference). The MISD optimization is raised mainly due to the \textit{compute scaling}, \ie, the tremendous computing performance of recent GPUs (like V100/A100s with $\sim$130TFLOPs/s) overwhelms the general DNNs' inference requirement (\eg, ResNet50 with $\sim$4GFLOPS).
Therefore, due to the practical cost consideration (\eg, cost of hardware, power consumption), data center-level serving also tends to adopt such a multi-tenant serving paradigm to effectively reduce the total cost of ownership (TCO).  

\vspace{1mm}
\noindent $\bullet$ {\textbf{Single Instance Multi Device (SIMD)}} SIMD optimization, to the contrary of MISD, is mainly raised by the \textit{model scaling} trend. The ultra-scale model size, especially in language~\cite{bert,gpt} and recommendation~\cite{dlrm} domains, has shown to be an important factor in improving the model accuracy. As a result, recent state-of-the-art industrial models have achieved the volume of hundreds of billions parameters~\cite{mega-turing} and the model inference can take $\sim$TB space in host memory~\cite{dlrm}, which is nearly impossible to be served on a single host. In such scenario, distributed inference by multiple devices becomes the only solution for such ultra-large model inference~\cite{fb}. Nevertheless, as slow inter-device communication channels are involved (\eg, remote process calls, RPCs), distributed inference requires both algorithm and scheduling innovations to achieve inference efficiency. For example, we need to consider both efficient model sharding in the algorithm side such as model/pipeline parallelism~\cite{pipelayer,gpipe}\footnote{Although such model/pipeline parallelism is mostly used for large model training, recent ultra-large model inference also requires such model sharding techniques so as to run on several devices.} and the inter-node co-scheduling in the scheduling side.

\vspace{1mm}
\noindent $\bullet$ {\textbf{Multi Instance Multi Device (MIMD)}} Towards even complexer scenarios, Multi-Instance Multi-Device (MIMD) optimizations consider how to route various model inference requests to different devices (\eg, service routers) and manage compute device utilization. Such optimization mainly lie in data center-level management for optimal infrastructure utilization and cost. Currently, public available works~\cite{fb_center,msr_center,antman} mainly target at optimizing training jobs consuming more resources (\eg, 4/8-GPU machines, taking hours to days). There are still limited public works targeting at inference MIMD optimizations. 

\begin{figure*}[!t]
	\centering
	\includegraphics[width=7in]{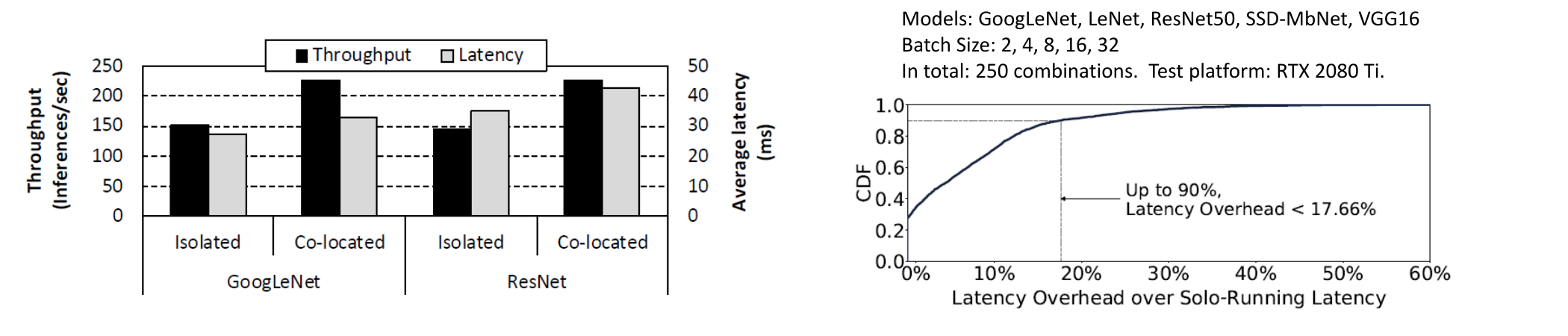}
	\vspace{-5mm}
	\caption{Multi-tenant inference latency and throughput comparison. (a) Although co-locating multiple DNNs degrades the latency by 5\%-10\%, the overall throughput can be improved 25\%+ on average~\cite{prema}. (b) Among 250 model co-location combination experiments, \eg, up to 90-percent of bi-model execution shows less than 17\% latency degradation~\cite{kaist}.}
	\vspace{-2mm}
	\label{fig:multi_tenant}
\end{figure*}


\section{{Computing Scaling: Trending to \\ Multi-Instance Single-Device (MISD)}}
\label{sec:misd}

\vspace{1mm}
\subsection{Overview} 

With the compute scaling, the capacity of recent hardwares has achieved exponential growing for deep learning. Especially for GPUs, recent GPUs have demonstrated overwhelming capacities (\eg, $\sim$130 TFLOPs/s for A100) compared to common model inference workload (\eg, ResNet50 with $\sim$4 GFLOPs).
As executing single DNN instance on such hardwares can incur severe resource under-utilization, multi-instance single-device (MISD) computing paradigm, or multi-tenant inference, co-locates multiple DNN instances onto the same hardware device to improve the hardware utilization as well as the serving throughput.




\vspace{1mm}
\noindent \textbf{Enhancing Serving Throughput}
One of the major goal of MISD is to achiever higher {serving throughput}, which is usually measured by the served queries per second (QPS).
%
By improving the resource utilization (\eg, computing units, memory bandwidth), MISD could increase the serving throughput on the same hardware.
Figure~\ref{fig:multi_tenant} (left) shows an example of co-running two DNN models (GoogLeNet and ResNet) on the same hardware. Although the latency of each model degrades by 5\%-10\%, the overall throughput is improved by 25\%+ on average. 

\begin{figure}[!b]
	\centering
	\vspace{-2mm}
	\includegraphics[width=3.4in]{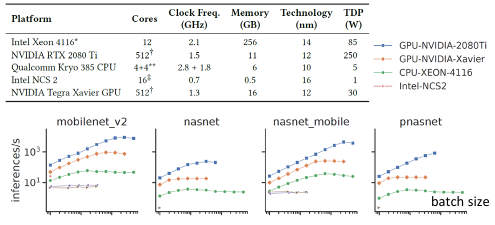}
	\vspace{-2mm}
	\caption{Throughput and Power Comparison of CPUs/GPUs~\cite{embench}.}
	\label{fig:power}
\end{figure}

\vspace{1mm}
\noindent \textbf{Reducing Power and Infrastructural Costs}
By enhancing the serving throughput, another benefit of MISD is it could lower both the cost of infrastructures, as well as reduce the average power needed for processing each query. 
Figure~\ref{fig:power} compares the power and serving throughput between CPU-based and GPU-based serving. Although the RTX2080Ti has 3$\times$ power consumption compared to Intel Xeon416 CPU (250W vs. 85W), the GPU serving throughput can reach at most 100$\times$, \eg, for MobileNetV2 and NasNet. This translates to  $\sim$30$\times$ average power reduction for processing each query, thus greatly reducing the power and related infrastructural costs. 

\vspace{1mm}
\noindent \textbf{Latency Performance Degradation}
However, co-locating more DNN instances would have worse latency performance since the average available resources are less for each DNN instance. 
Therefore, DL inference service providers and consumers will usually set certain latency constraints (SLA), which requires queries to be served within given latency (for example, less than 100ms for ads display for the best user experience).
Thus, multi-tenant inference with certain latency degradation in SLA range is also considered acceptable.
Figure~\ref{fig:multi_tenant} (right) shows an example of the latency degradation analysis among 250 pairs of model co-running combinations, the average latency degradation is only 17\% for up to 90\% of the combinations, demonstrating the great potential of GPU utilization enhancement. 

\begin{table*}[!t]
\scriptsize
\caption{Recent Works on Multi-Tenant Inference Optimization (JCT: job completion time, SLA: service-level agreement).}
\renewcommand\arraystretch{1.5}
\begin{tabular}{llllll}
\hline
\multicolumn{1}{c}{\textbf{Ref.}} &
  \multicolumn{1}{c}{\textbf{Hardware}} &
  \multicolumn{1}{c}{\textbf{Problem}} &
  \multicolumn{1}{c}{\textbf{Perspective}} &
  \multicolumn{1}{c}{\textbf{Algorithm/Strategy}} &
  \multicolumn{1}{c}{\textbf{Improvement/Achievement}} \\ \hline
\cite{iccad} &
  GPU &
  \begin{tabular}[c]{@{}l@{}}$\bullet$ Resource under-utilization\\$\bullet$ Contention\end{tabular} &
  SW Scheduling &
  \begin{tabular}[c]{@{}l@{}}$\bullet$ Operator-level scheduling\\$\bullet$ ML-based scheduling auto-search\end{tabular} &
  $\bullet$ Reduced inference makespan \\ \hline
\cite{stanf} &
  GPU &
  $\bullet$ Inter-job interference &
  SW Scheduling &
  \begin{tabular}[c]{@{}l@{}}$\bullet$ Query-level online scheduling\\ $\bullet$ ML-based interference predictor\end{tabular} &
  $\bullet$ Reduced latency \\ \hline
\cite{irina} &
  GPU &
  $\bullet$ Client-side waiting time &
  SW Scheduling &
  \begin{tabular}[c]{@{}l@{}}$\bullet$ Query-level online scheduling\\$\bullet$ Heuristic-based preemption\\$\bullet$ Concurrent and batching\end{tabular} &
  $\bullet$ Reduced latency \\ \hline
\cite{prema} &
  NPU &
  $\bullet$ Priority-based serving &
  SW Scheduling &
  \begin{tabular}[c]{@{}l@{}}$\bullet$ Query-level online scheduling\\$\bullet$ Heuristic preemption\end{tabular} &
  \begin{tabular}[c]{@{}l@{}}$\bullet$ Reduced high-priority job JCT\\$\bullet$ Maintaining low-priority SLA\end{tabular} \\ \hline
\cite{gslice} &
  GPU &
  \begin{tabular}[c]{@{}l@{}}$\bullet$ Resource under-utilization\\$\bullet$ contention\end{tabular} &
  HW Resource Managing &
  \begin{tabular}[c]{@{}l@{}}$\bullet$ Managed resource provisioning\\$\bullet$ Adaptive batching\end{tabular} &
  \begin{tabular}[c]{@{}l@{}}$\bullet$ Enhanced serving throughput\\$\bullet$ Maintaining SLA\end{tabular} \\ \hline
\cite{kaist} &
  GPU &
  \begin{tabular}[c]{@{}l@{}}$\bullet$ Resource under-utilization\\$\bullet$ Contention\end{tabular} &
  HW Resource Managing &
  \begin{tabular}[c]{@{}l@{}}$\bullet$ Managed resource provisioning\\$\bullet$ Adaptive batching\end{tabular} &
  \begin{tabular}[c]{@{}l@{}}$\bullet$ Enhanced serving throughput\\$\bullet$ Maintaining SLA\end{tabular} \\ \hline
\cite{plana} &
  \begin{tabular}[c]{@{}l@{}}Systolic\\ Arrays\end{tabular} &
  $\bullet$ Resource under-utilization &
  Architecture ReConfig &
  $\bullet$ Hardware resource fission &
  \begin{tabular}[c]{@{}l@{}}$\bullet$ Enhanced serving throughput\\$\bullet$ Reduced energy cost\end{tabular} \\ \hline
\end{tabular}
\label{table:misd}
\end{table*}
\normalsize

\subsection{Challenges in MISD}
\label{sec:misd_prob}

There are many challenges that hinder the performance improvement of MISD. We summarize the major ones, such as inter-tenant interference and serving workload dynamics.

\vspace{1mm}
\subsubsection{Inter-Tenant Interference}

In MISD, co-locating multiple DL jobs on the same hardware allows these models to share the compute resource, while brings the problem of inter-tenant interference. 
As DNN models contain many types of operators such as convolution, batch-norm, skip-connection, \etc, different types of operators can be either compute intensive (\eg, convolution) or memory intensive (\eg, skip-connection). 
Therefore, co-running multiple models on the GPU can incur resource contention in computing or memory bandwidth due to the poor operator concurrency management (\eg, co-running the same type of operators at the same time). 
As a result, the inference speed of both models can be slowed down.
Such cases can happen frequently with more increased number of tenants and thus degrade the overall serving throughput.

\subsubsection{{Serving Workload Dynamics}}
The inter-tenant interference could be more difficult to handle when the serving workload is not static but dynamic.
For example, in the cloud environment, due to the fact that DL jobs' arrival is usually non-predictable, perfectly interleaving compute-intensive and memory-intensive DNN queries is usually not feasible. 
In such case, multi-tenant inference and potential interference behavior is hard to predict as it has a tremendous combination space due to various of models, different number of co-running jobs, \etc 
With such unpredictability, it is very challenging to maintain no inter-tenant interference and achieve ideal serving throughput.

\subsubsection{{Job Priority and Completion Time}} 
There are several other problems in MISD computing.
For example, one important issue that needs to be considered is the job priority, \eg, certain jobs need to be finished with ensured performance or with higher priority. In such case, preemption is a common scheduling strategy to ensure high-priority jobs to finish in a short time. Or we can also allocate dedicated resource to high-priority jobs.
Another issue is that instead of considering server-side serving throughput, certain serving systems also need to consider client-side user experiences measured by average job completion time (JCT). Minimizing JCT is also contradictory to maximizing throughput, \eg, concurrently running multiple jobs can cause longer average JCT but increase serving throughput.

\subsection{Optimization Techniques}
\label{sec:misd_tech}

We summarize recent works in multi-tenant inference optimization into two major categories: temporal workload scheduling and spatial resource management. An overview of these works are shown in Table~\ref{table:misd}.


\begin{figure}[!b]
	\centering
	\vspace{-2mm}
	\includegraphics[width=3.4in]{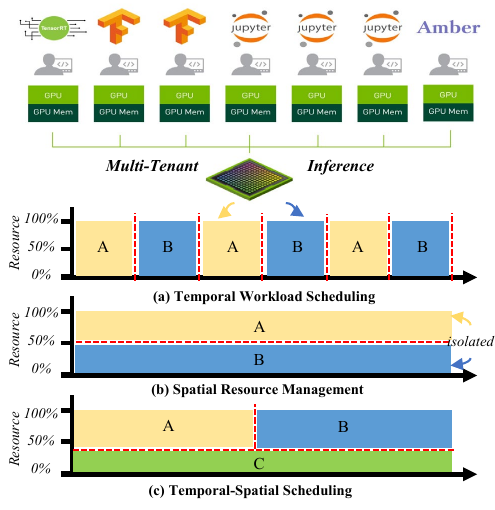}
	\vspace{-2mm}
	\caption{Multi-Tenant Inference with Temporal and Spatial Scheduling.}
	\label{fig:misd_overview}
\end{figure}

\subsubsection{Software-Level Workload Scheduling}

As we mentioned before, one of the major challenges in multi-tenant inference is to avoid job interference.
Therefore, many research works~\cite{iccad,stanf,irina, prema} conduct workload scheduling among different DL jobs temporally to reduce such interfering, as shown in Figure~\ref{fig:misd_overview}. 
Such solutions aims to avoid interference by scheduling different DL model workloads along the time dimension, thus we categorize such works as software-level scheduling solutions.

In practice, such workload scheduling granularity could be fine-grained, \eg, scheduling the DNN operators, or coarse-grained, \eg, scheduling the entire inference query. 

\vspace{1mm}
\noindent \textbf{Operator Scheduling} 
To achieve fine-grained scheduling, works~\cite{iccad, ios} regard the DL model operators (\eg, conv, batch-norm, identity) as the minimum scheduling units.
They first abstract multiple DNN's computation graph with many operators into an intermediate representation (IR). 
To explore the huge scheduling space, they design a ML-based auto-search method by defining three main factors: scheduling search space, profiling-guided latency cost model, and the machine learning-based search algorithm to find the best scheduling for a balanced GPU utilization without interference and reduced latency. Such fine-grained operator scheduling could achieve better performance, but usually face scalability issues when the number of workloads increases to very large, such as hundreds of DNN queries.

\vspace{1mm}
\noindent \textbf{Service Router} 
Therefore, other works like~\cite{stanf,irina,prema} use a more coarse scheduling granularity, \eg, regarding each query as the minimum scheduling units, which thus reduces the scheduling complexity. 
For example, one of the  query-level workload scheduling example is the service router in large-scale data center, such as Microsoft Deep Learning Inference Service (DLIS) system~\cite{dlis}. To achieve high utilization while avoid resource contention, service router needs to understand different models’ requirements and place one or multiple queries intelligently onto hardware. 

\vspace{1mm}
\noindent \textbf{Predictive Scheduling} 
Within above scheduling techniques, one challenging factor is the {serving queries dynamics}, for example, unknown number of incoming queries, RNN/LSTMs with varied input lengths. Different from static workloads that we can get the full workload information for scheduling, such dynamic workloads require us to consider the potential incoming workloads. To handle such dynamics, PREMA~\cite{prema} proposed a predictive multi-task DNN scheduling algorithm that combines off-line profiling latency records and an online token-based job scheduling algorithm. Meanwhile, it also enables adaptive job preemption to achieve priority control. But as each DNN can have many operators (\eg, layers) that have fluctuated resource consumption, such coarse-grained scheduling may still suffer occasionally resource under-utilization/contention and sub-optimal performance.

\subsubsection{Hardware-Level Resource Management}

Besides temporal scheduling, another optimization perspective to solve the inter-tenant inference is to conduct fine-grained resource managing~\cite{gslice, plana}, such as spatial partitioning and isolation of resources for different jobs, which we consider as a spatial perspective. As such partitioning could isolate different job's used resource (\eg, stream multiprocessors (SMs), memory bandwidths), such solution can help avoid the job interference in the hardware level.

\vspace{1mm}
\noindent \textbf{Resource Partitioning}
Previously, achieving fine-grained resource partitioning is non-achievable until recently NVIDIA GPUs release a series of resource sharing and partitioning support like multi-streams (MST), multi-process services (MPS~\cite{mps}) and multi-instance GPU (MIG~\cite{mig})\footnote{The major differences between three tools are that MST only enables resource sharing but doesn't avoid SM nor memory bandwidth interference. MPS enables SM partitioning by allocating certain SMs to given jobs (\eg, for 5 DNNs, allocating 20\% SMs for each one), thus avoiding SM interference but doesn't address memory interference. MIG achieves both SM and memory bandwidth partitioning, which is recently supported on NVIDIA A100 GPUs only~\cite{mig}.}.
Leveraging such support, \cite{gslice} uses MPS to conduct adaptive SM partitioning for different DNN jobs. By doing so, the SM resource contention could be avoided among co-located DL workloads.
Similarly, \cite{plana} utilizes special accelerator (systolic arrays) and implements architecture support for hardware fission, which achieves fine-grained resource managing. 

\vspace{1mm}
\noindent \textbf{Hardware Re-Configuration}
However, such spatial resource partitioning solutions also have a intrinsic limitation that is the inflexible re-configuration when facing dynamic workloads. For both GPUs and other accelerators, changing the resource partitioning configurations requires certain amount of time (\eg, tens of seconds or more), which can be much larger than DL workloads' processing time (usually served in ms). Therefore, re-configuring the resource partitioning frequently is usually not practical and thus limits such solutions' performance when facing dynamic workloads. \cite{gslice} tries to reduce the stall caused by reconfiguration time of MPS by utilizing a standby/shadow process. However, the minimum time for switching one partitioning configuration to another one still cost several seconds, which is non-negligible in online serving scenarios.

\subsection{Future Directions}
\label{sec:misd_future}

\vspace{1mm}
\subsubsection{Software-Hardware Co-Scheduling}

The software and hardware scheduling could be complementary to each other to provide both high job scheduling flexibility and strict resource isolation. Recently, there are certain works that adopt a temporal-spatial combined perspective to achieve better serving performance. \cite{kaist} adopts MPS to conduct resource partitioning and split one GPU to multiple gpulets and then implements a heuristic-based greedy task scheduler to find the appropriate mapping relationship between the current DNN queries and gpulets. 
Nevertheless, the serving workload dynamics in spatial-temporal scheduling is more complex and needs to be re-thinked and designed. For example, beyond previous temporal-only scheduling that only decides when to start each job, having multiple hardware partitions also requires to decide which one to place on. The current greedy task scheduler in~\cite{kaist} treats the current workload as static without considering potentially dynamic incoming job requests. Thus, a simple sub-optimal case can be that it can allocate a current small workload to a large gpulet, while leaving no gpulet for a larger incoming DNN job to use, resulting in potential resource under-utilization.

\subsubsection{ML-based Prediction and Online Learning}

To avoid such problem, we can potentially use a ML-based predictive model (\eg, reinforcement learning, LSTM, \etc) to predict the dynamic workload and then guide the workload-to-partition mapping. The ML-based model can be initially trained offline by historical serving records. During the online serving process, active and lifelong learning, \ie, using the latency/throughput as feedback to consistently improve the predictive accuracy, can also be potentially utilized in such case to improve the workload prediction and scheduling effectiveness.


Another way of leveraging ML-based prediction is to conduct modeling to predict the potential latency performance under different multi-model and hardware combinations so that the scheduler can make better decision regarding the latency SLA constraints. For example, the work~\cite{stanf} built a ML model to predict the latency of multi-model inference cases on different machines. However, the effectiveness of such solution also highly depends on the modeling accuracy, scalability and generality, which can be hard to achieve all in practice.


\section{Model Scaling: Trending to \\ Single Instance Multi Device (SIMD)}
\label{sec:simd}

\vspace{1mm}
\subsection{Overview}

Different from MISD, the trend of SIMD is mainly brought by the tremendous model scaling~\cite{gpt,mega-turing} which makes one model hard to execute on one single machine.
For example, industrial recommendation models~\cite{dlrm} requires $\sim$TB of memory for inference.
In such cases, scaling up single machine to support such memory capacity can incur huge infrastructural cost in large-scale data centers. Therefore, one more cost-efficient way is to scale out, which is to using multiple distributed compute devices to conduct collaboratively inference.
However, unlike distributed training with long time duration and thus loose time constraints, inference serving have strict latency constraints. Thus, the communication cost in such collaboratively serving can make the SLAs more challenging to achieve. 

\vspace{1mm}
\noindent \textbf{{Enhancing Capacity Scaling}} 
The first priority of SIMD is to achieve capacity scaling of infrastructures so as to enable super-scale model inference serving. 
To do so, there are two major ways: \textbf{Scale up vs. Scale out}. 
Scale up (or vertical scaling) means upgrading or adding more resources to an system to reach a desired state of performance, \eg, more compute, memory, storage and network bandwidth, \etc
By contrast, scale out (or horizontal scaling) uses more distributed low-end machines to collaboratively serve one large DL model. SIMD such as distributed inference~\cite{fb,deepthing,hotedge,coedge} mainly uses the second scale-out solution to maintain the capacity scaling for larger DL model serving. Recently, heterogeneous inference~\cite{hm1,hm2,hm3} comes up, which combines heterogeneous resources such as hybriding CPU/GPU, hybriding cache/memory/SSD, etc. Such solutions could be considered as a scale-up solution. 

\vspace{1mm}
\noindent \textbf{{Reducing Costs}} 
For both scaling up and scaling out solutions, the major optimization targets of above solutions (\ie, SIMD) is to reduce the total cost of ownership (TCO).
For example, scaling up by increasing single machine capacity such as upgrading CPU generations, increasing memory capacity and frequencies can be expensive within large-scale data center-level infrastructures. 
Therefore, the heterogeneous inference mainly relies on existing hardwares and combines them to enhance the compute capacity, thus reducing the capacity scaling costs.
By contrast, scaling out utilizes multiple current generation of compute devices to serve next-generation larger models, thus also reducing the cost of setting up new dedicated hardware.
For both methods, the shared downside is that it brings complex distributed computing and communication management, and even potential performance degradation.

\begin{figure}[!b]
	\centering
	\includegraphics[width=3.4in]{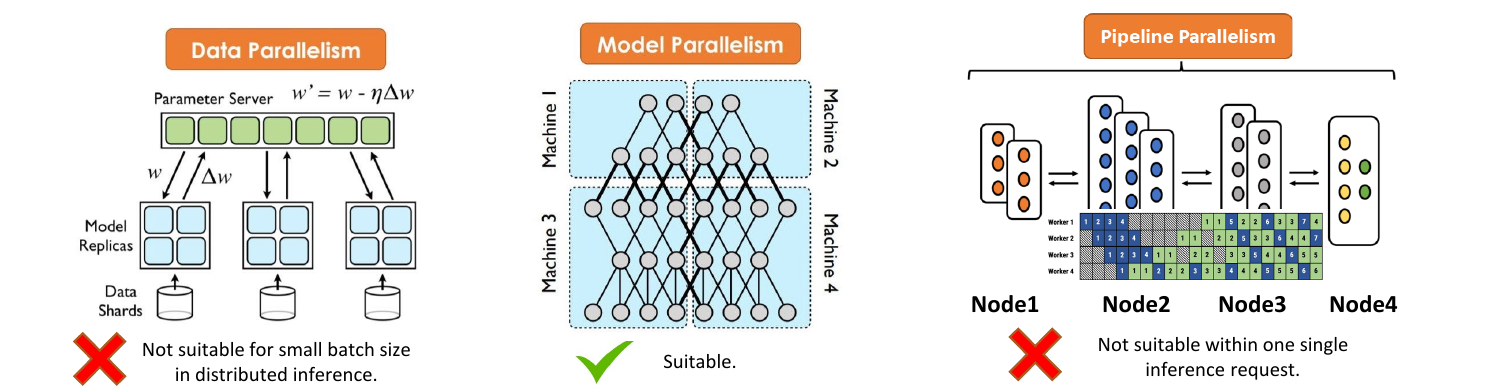}
	\caption{In three parallelisms, model parallelism is the common way to achieve distributed inference while the other two are only suitable for distributed training. For example, data parallelism is not suitable for inference with small batch sizes. Pipeline parallelism cannot leverage multi-node computing parallelism within a single inference request.}
	\label{fig:paral}
\end{figure}

\subsection{Challenges in SIMD}

\vspace{1mm}
\subsubsection{{Computing Parallelism}} 

Although SIMD with distributed devices can achieve the model's required overall capacity, distributing one entire model to collaboratively compute on multi-nodes is still very complex. 
There are many model partitioning algorithms that have been explored in {distributed training} algorithms such as data, model, and pipeline parallelism~\cite{pipelayer, pipedream, hidden}, \etc  
However, not all of them are suitable for {inference} scenarios.
As shown in Figure~\ref{fig:paral}, traditional data parallelism by splitting the batch sizes is usually not applicable since the inference serving usually comes with small batch sizes. 
Pipeline parallelism~\cite{pipelayer,pipedream,gpipe} by allocating different layers into different machines is also not suitable as it cannot achieve multi-node computing parallelism within one single inference request. 
As a result, model parallelism~\cite{hidden} is the most common way to achieve SIMD or distributed inference, as we will introduce later.

\subsubsection{{Communication Bottlenecks}} 
For model parallelism on multi-distributed machines, achieving linear or even sub-linear speedup is usually very hard. The reason is that, the major factor that influences the computing efficiency (such as latency) in model parallelism is the communication bottlenecks caused by the intermediate data transfer between different model shards (distributed nodes)~\cite{gpipe,commu1}. Depending on the data transmission speed, too much data transfer will incur significant latency overhead, \eg, with Ethernet connections in distributed machines. To avoid that, an efficient model sharding algorithm is needed to ensure the inference serving performance in MISD.
However, the intermediate data communication mechanisms vary a lot in different model architectures, such as CNNs, Transformers, and recommendation DLRMs, \etc, and thus it requires many analysis efforts and design innovations to find the optimal model sharding algorithm.

\begin{figure*}[!t]
	\centering
	\includegraphics[width=6in]{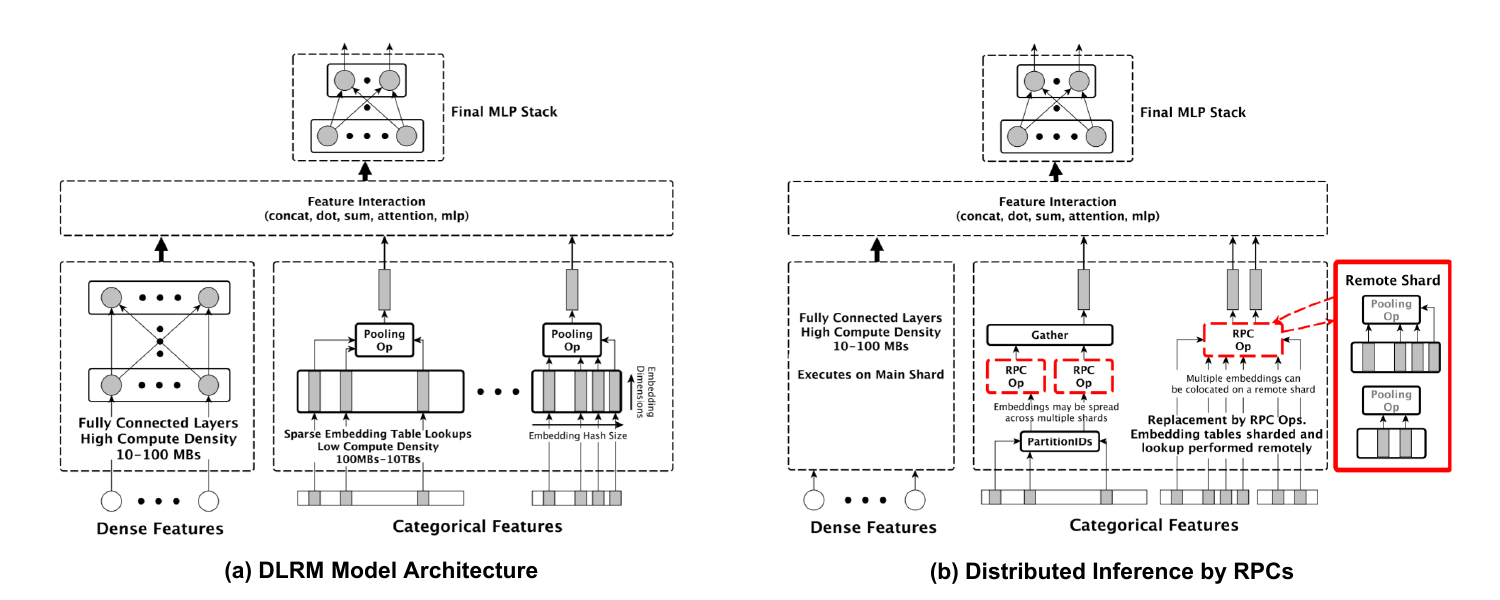}
	\vspace{-2mm}
	\caption{Distributed Inference could use RPCs to leverage multiple distributed hosts to collaboratively execute one large recommendation model~\cite{fb}.}
	\vspace{-2mm}
	\label{fig:fb}
\end{figure*}

\subsection{Optimization Techniques}

We mainly introduce two aspects of works that conduct SIMD: Scaling out by distributed inference and Scaling up by heterogeneous inference. 

\vspace{1mm}
\subsubsection{{Scaling Out: Distributed Inference}}

\vspace{1mm}
\noindent \textbf{Recommendation Model Distributed Inference}
For example, recent industry-level deep learning recommendation model (DLRM)~\cite{dlrm} has reached $\sim$TB-level model size and such recommendation request takes over 79\% of Facebook's data center's workload, \cite{fb} proposed the first work that applies the distributed inference technique into the large DLRM inference serving. Different from conventional MLPs and CNNs, DLRMs are mainly composed of \eph{feature embedding tables} and fully-connected layers, where feature embedding tables are the major part of model weights (up to 80\% to 95\%). Such feature embedding part of DLRMs is memory-intensive, but have very light computation workload as it only needs to access certain columns of the embedding tables and conducts summation~\cite{fb}. 
%

Therefore, distributed inference is proposed to partition large embedding tables into different compute nodes and thus overcome the inference serving memory limitation. When the table embeddings in remote nodes are required, central model could invoke remote process calls (RPCs) to directly get the computed results from remote nodes and thus fulfill the distributed inference serving.


\vspace{1mm}
\noindent \textbf{CNN Distributed Inference}
Compared to DLRMs, CNN model parallelism is more widely studied (such as channel parallelism/spatial parallelism)~\cite{hotedge} and many works have also propose the CNN models' distributed inference~\cite{coedge,deepthing}. For example, CoEdge~\cite{coedge} is one example of cooperative CNN inference with adaptive workload partitioning over heterogeneous distributed devices. Similarly, DeepThings~\cite{deepthing} also proposes a framework for adaptively distributed execution of CNN-based inference on resource constrained IoT edge devices. It designs a scalable Fused Tile Partitioning (FTP) to minimize memory footprint while exposing parallelism to collaboratively compute of major convolutional layers. It further realizes a distributed work stealing approach to enable dynamic workload distribution and balancing at inference runtime.

\subsubsection{{Scaling Up: Heterogeneous Inference}}

\vspace{1mm}
\noindent \textbf{Heterogeneous Memory} Besides using distributed machines, another way to address super-scale model's large memory capacity requirement is to leverage heterogeneous memory, \eg, by combining different levels of memory such as cache, main memory and even SSDs. The major challenge of such heterogeneous memory is to address the slow access speed of SSD (the access bandwidth can be usually 100x slower than memory). Targeting at this bottleneck, \cite{hm1,hm2,hm3} have proposed similar heterogeneous memory design that leverages storage-level SSDs to store partial of the embedding table weights of the DLRM. As the embedding table access patterns of DLRMs are usually sparse and have certain spatial/temporal locality, a proper embedding table placement strategy between memory/SSD with dedicated caching strategy could thus greatly enhance the heterogeneous memory access efficiency, thus reaching on-pair performance with pure memory based performance.

\vspace{1mm}
\noindent \textbf{Heterogeneous Computing} Along with heterogeneous memory, heterogeneous computing is a similar concept by hybriding the computing capacity for different processors, which is mostly commonly used for large model inference in SoCs with multiple types of computing processors such as CPU, GPU and other accelerators. For example, Synergy~\cite{synergy} proposed a HW/SW framework for high throughput CNNs on embedded heterogeneous SoC. It leverages all the available on-chip resources, which includes the dual-core ARM processor along with the FPGA and the accelerator to collaboratively compute for the CNN inference. Similar to distributed inference which partition one model onto multiple machines, heterogeneous computing also requires dedicated model sharding algorithms and co-designs the model sharding with the heterogeneous computing devices, and also considers the inter-processor communication, \etc




\section{Conclusion}
\label{sec:con}

In this survey, we propose a a novel computing paradigm taxonomy to characterize ongoing large-scale deep learning system (LDS) optimization works.
We summarize the new optimization objectives, elaborate new technical design perspectives, and provide the insights for future LDS optimization.
By summarizing the existing works, we hope that this survey could provide a comprehensive summary on emerging challenges, opportunities, and innovations, and thus motivates new innovations in LDS operation and optimization. 

\bibliographystyle{ACM-Reference-Format}
\bibliography{sample-base}

\end{document}